\documentclass[runningheads]{llncs}

\usepackage{eccvabbrv}

\usepackage{graphicx}
\usepackage{booktabs}

\usepackage[accsupp]{axessibility}  

\usepackage{textcomp}

\usepackage{hyperref}

\usepackage{orcidlink}

\begin{document}

\title{EaDeblur-GS: Event assisted 3D Deblur Reconstruction with Gaussian Splatting} 

\titlerunning{EaDeblur-GS}

\author{Yuchen Weng\orcidlink{0009--0005--4219--3108} \and
Zhengwen Shen\orcidlink{0000--0001--8882--5960}\and
Ruofan Chen\orcidlink{0009--0002--0382--7479}\and
Qi Wang\orcidlink{0009--0003--1553--0493}\and
Shaoze You\orcidlink{0000--0002--0488--0012}\and
Jun Wang \thanks{Corresponding author: \email{jrobot@126.com}} \orcidlink{0000--0002--8926--156X}}

\authorrunning{Y.Weng et al.}

\institute{School of Information and Control Engineering, China University of Mining and Technology, 221000 Xuzhou, China}

\maketitle

\begin{abstract}
  3D deblurring reconstruction techniques have recently seen significant advancements with the development of Neural Radiance Fields (NeRF) and 3D Gaussian Splatting (3DGS).
  Although these techniques can recover relatively clear 3D reconstructions from blurry image inputs, they still face limitations in handling severe blurring and complex camera motion.
  To address these issues, we propose Event-assisted 3D Deblur Reconstruction with Gaussian Splatting (EaDeblur-GS), which integrates event camera data to enhance the robustness of 3DGS against motion blur.
  EaDeblur-GS utilizes a novel Adaptive Deviation Estimator (ADE) network and two novel loss functions to achieve real-time, sharp 3D reconstructions.
  Evaluations demonstrates that our method achieves advanced performance against original Gaussian Splatting and other Deblur Gaussian Splatting techniques.
\end{abstract}

\section{Introduction}
\label{sec:introduction}
Reconstructing 3D scenes and objects from images has long been a research hotspot in computer vision and computer graphics. 
The advent of Neural Radiance Fields (NeRF) \cite{mildenhall2021nerf} has brought revolutionary advancements in photo-realistic novel view synthesis.
Building on this progress, the recent introduction of 3D Gaussian Splatting (3DGS) \cite{kerbl20233d} has further enhanced 3D scene representation using Gaussian ellipsoids, achieving high-quality 2D rendering and faster training and rendering speeds.
However, in real-world applications, factors like camera shake and shutter speed often lead to image blurriness and inaccurate camera poses estimation, challenging the clear neural volumetric representation. 

Several methods address blurriness in NeRFs and 3DGS.
NeRF's deblurring technology has undergone relatively early development, with Deblur-NeRF \cite{ma2022deblur} being the first framework to tackle this issue, which utilized an analysis-by-synthesis approach to recover sharp NeRFs from blurry inputs.
MP-NeRF\cite{wang2024mp} further enhances this by introducing a multi-branch fusion network and prior-based learnable weights to handle extremely blurry or unevenly blurred images.
But the NeRF-based methods always consume extensive training time and rendering time.
Hence, some methods based on 3DGS has been developed because of its advantages in rendering and training speed.
For instance, Wenbo Chen et al.~\cite{Chen_deblurgs2024} proposed Deblur-GS, which models the problem of motion blur as a joint optimization involving camera trajectory and time sampling.
B. Lee etal.~\cite{lee2024deblurring} assign the corrections on the rotation and scaling matrix of 3D gaussians by using a small MLP, enhancing the clarity of scenes reconstructed from blurred images.
Nonetheless, these methods can only achieve clear 3D reconstruction results with mildly blurred input images.
Consequently, additional data sources like event cameras have been introduced into 3D deblurring reconstruction.
Event cameras, a bio-inspired sensor, offers high temporal resolution and has advantages in motion deblur.
For NeRFs, EventNeRF \cite{rudnev2023eventnerf} used event integration and color simulation for colored 3D representations, while Qi et al. \cite{qi2023e2nerf} combined blurred images with event streams using innovative loss functions and the Event Double Integral (EDI) approach.
For 3DGS, Yu proposed EvaGaussian\cite{yu2024evagaussians}, optimizing both scene reconstruction and camera trajectory estimation to achieve unparalleled detail and fidelity in real-time 3D scene synthesis.

Despite achieving excellent performance in 3D deblurring reconstruction, the aforementioned methods still have limitations. 
For instance, RGB single-modality deblurring 3DGS and NeRF are often effective only in mildly blurred or simple camera motion scenarios.
When input images are severely blurred, these methods usually fail to reconstruct 3D objects and can only produce relatively clear 2D renderings from certain angles.
On the other hand, techniques using event cameras for NeRF 3D reconstruction are limited by NeRF's training and rendering speed. Additionally, methods that incorporate event data into 3DGS deblurring reconstruction face challenges, such as inaccurate camera motion trajectory estimation.


To overcome these limitations, we propose a novel integration of event streams with 3DGS, namely Event-assisted 3D Deblur Reconstruction with Gaussian Splatting(EaDeblur-GS), aiming to guide the learning of better 3D Gaussian representations and tackle issues arising from blurry inputs. 

Our contributions are summarized as follows:
\begin{itemize}
\item We propose Event-assisted 3D Deblur Gaussian Splatting (EaDeblur-GS), which incorporates blurry RGB images and event streams into Gaussian Splatting to recover sharp 3D representations.
\item We introduce a novel Adaptive Deviation Estimator (ADE) network to simulate the shaking motion during exposure accurately by estimating the deviations of Gaussians.
\item We comprehensively evaluate the proposed method and compare it with several baselines, demonstrating that our EaDeblur-GS achieves advanced performance while enabling real-time sharp image rendering.
\end{itemize}

\section{Method}
\label{sec:Method}
\begin{figure}
    \centering
    \includegraphics[width=1.0\textwidth]{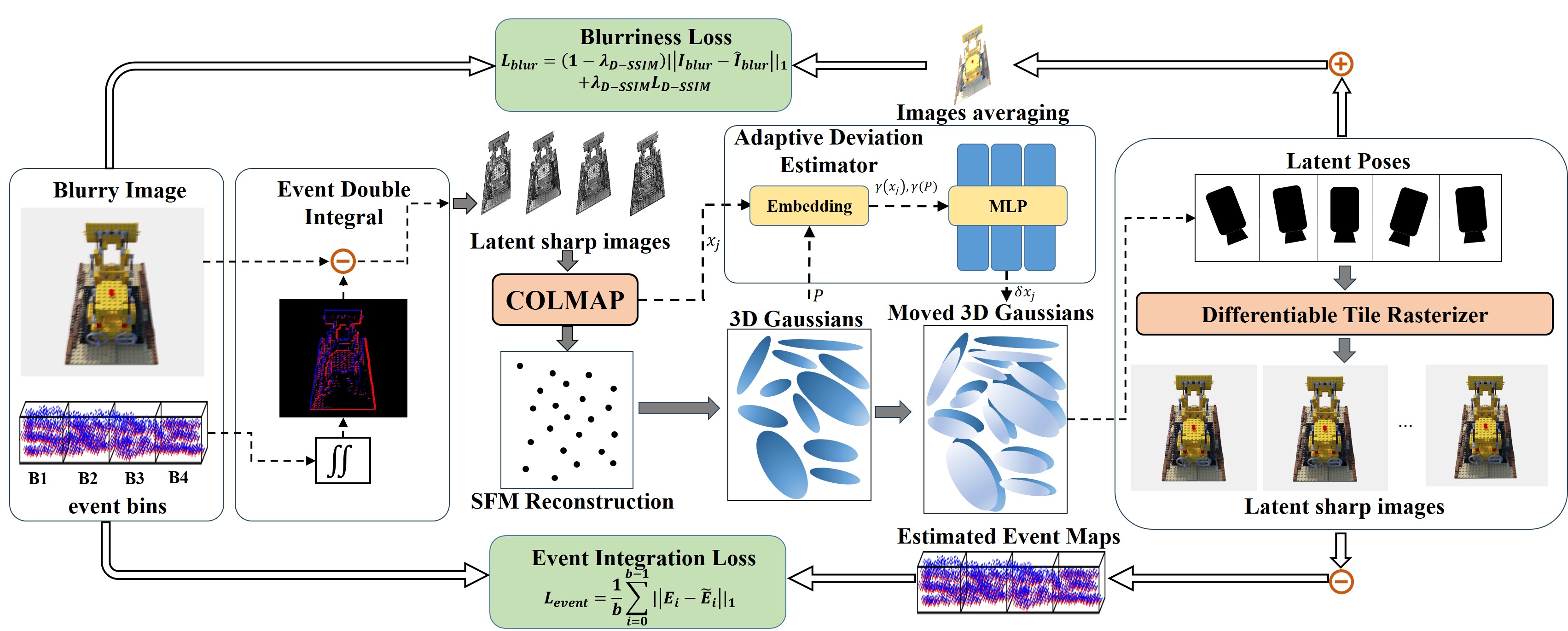}
    \caption{The Overview of our approach.
    Our approach integrates blurred RGB images with event data to enhance image clarity. 
    The EDI technique is applied to produce latent sharp images, which are then refined through COLMAP for accurate SFM reconstruction and 3D Gaussian modeling. 
    The ADE module estimates positional deviations based on initial Gaussian positions and camera extrinsics, simulating camera motion. 
    These deviated Gaussians are rendered into multiple views and compute blurriness and event integration losses, facilitating the learning of detailed 3D representations for enhanced reconstruction quality.   
    }
    \label{fig:EaDeblurGS}
\end{figure}

As illustrated in Fig.\ref{fig:EaDeblurGS}, our method begins with the input of blurry RGB images and their corresponding event streams. 
We first use the Event Double Integral (EDI) technique to generate a set of latent sharp images. 
These images are then processed with COLMAP, enhancing the initial reconstruction and providing relative precise camera pose estimation. 
From this reconstruction, we create a set of 3D Gaussians. 
These Gaussian positions, combined with the estimated camera poses, are then fed into our Adaptive Deviation Estimator (ADE) network to adjust the Gaussian centers by calculating positional deviations.
The adjusted 3D Gaussians are projected onto each viewpoint, including corresponding latent viewpoints, to render sharp images. 
We also introduce a Blurriness Loss to simulate realistic blurriness and an Event Integration Loss to improve object shape accuracy in the Gaussian model.
This process enables the model to learn precise 3D volume representations and achieve superior 3D reconstructions. The overall method is depicted in Fig.\ref{fig:EaDeblurGS}. 

In the following sections, we detail how the ADE network estimates deviations, followed by an in-depth introduction to blurriness loss and event integration loss.
\subsection{Adaptive Deviation Estimator}
Motion blur caused by camera shake impedes sparse initial reconstruction due to unclear input images. 
To mitigate this issue, we employ the EDI method \cite{pan2019bringing}, which combines blurry images and corresponding event streams. 
The EDI model transforms a blurry image $I_{blur}$ into multiple sharp images ${I_0,\cdots,I_k}$, under the assumption that the blurry image is the temporal average of latent sharp images, each represented by the accumulated events. 
Given a blurry image and its corresponding event bins $\{B_k\}_{k=1}^b$, the sharp image of viewpoint $I_0$ and each latent sharp image $I_k$ can be expressed as:
\begin{equation}
I_0=\frac{(b+1)I_{blur}}{1+e^{\Theta\sum_{i=1}^1B_i}+\cdots+e^{\Theta\sum_{i=1}^bB_i}}.
\label{eq:start_image}
\end{equation}
\begin{equation}
I_k=\frac{(b+1)I_{blur}e^{\Theta\sum_{i=1}^kB_i}}{1+e^{\Theta\sum_{i=1}^1B_i}+\cdots+e^{\Theta\sum_{i=1}^bB_i}}.
\label{eq:sharp_images}
\end{equation}
We then utilize COLMAP to estimate the poses of EDI-processed sharp images $(\{\mathbf{P}_k\}_{k=0}^b=\mathrm{COLMAP}(\{I_k\}_{k=0}^b)$, which also results in a more accurate initial sparse point cloud. 

Inspired by \cite{Chen_deblurgs2024} and \cite{lee2024deblurring}, we represent latent poses as displacements of Gaussian ellipsoid centers $x_{j}$.
To estimate the deviations of Gaussians in this context, we employ Adaptive Deviation Estimator network, which is consisted of a small Multi-Layer Perceptron (MLP) $\mathcal{F}_{\theta}$ with three hidden linear layers. 
Besides, we utilize an embedding layer which encodes low-frequency positional information into high-frequency expressions and a decode layer to output the estimated deviations.
The ADE $\mathcal{F}_{\theta}$ takes the EDI-predicted poses ${\mathbf{P}}_{k=0}^b$ and original positions of Gaussians $x_{j}$ to estimate deviations:
\begin{equation}
    \begin{aligned}\{(\delta x_j^{(i)}\}_{i=1}^l=\mathcal{F}_\theta\Big(\gamma(x_j),\gamma(P)\Big)\end{aligned}
\end{equation}
where $l$ means the number of estimated latent poses, and $\delta x_j^{(i)}$ denotes the $i$-th predicted position offset of $j$-th Gaussian.
We then obtain extra $l$ sets of 3D Gaussians $\{\{\hat{x}_j^{(i)}\}_{i=1}^l\}_{j=1}^{N_G}$ by adjusting the positions of the original 3D Gaussians, where $N_G$ is the number of Gaussians, $\hat{x}_j^{(i)}$ represents the position offset scaled by $\lambda_p:\hat{x}_j^{(i)}=x_j+\lambda_p\delta x_j^{(i)}$.
This method computes $l$ different sets of 3D Gaussians for each viewpoint, which can be rasterized to $l$ sharp latent images $\{I_i\}_{i=0}^{l}$.
Additionally, we rasterize an image with original Gaussians positions as the image rendered from original viewpoint.
During forward rendering process, the ADE network and multiple rendering are not necessary, ensuring real-time inference speed comparable to original 3D Gaussian. 

\subsection{Loss Functions}
\textit{\textbf{Blurriness loss.}} To model the motion blur process during exposure time, we take the average sum of rendered images as follows:
\begin{equation}
    \hat{I}_{blur}=\frac1{b+1}\sum_{i=0}^{b}I_i,\quad I_i=\text{Rasterize}(\{G(\hat{x}_j^{(i)})\}_{j=1}^{N_G})
\end{equation}
where $I_i$ is the sharp images generated by multiple rendering,  $\hat{I}_{blur}$ is the estimated blurry image.
The blurriness loss is computed as the difference between estimated blurry image and input blurry image $I_{blur}$, combined with a D-SSIM loss as follows:
\begin{equation}
    L_{blur}=(1-\lambda_{D-SSIM})\|I_{blur}-\hat{I}_{blur}\|_1+\lambda_{D-SSIM} L_{D-SSIM}
\label{eq:blurloss}
\end{equation}
where we set $\lambda_{D-SSIM}=0.2$ for all experiments.

\noindent\textit{\textbf{Event Integration Loss.}} 
Leveraging the high time-resolution event stream, we adopt an Event Integration Loss to guide the network in learning a fine-grained sharp 3D reconstructions.
We integrate the events polarities between two input blurry frames as follows:
\begin{equation}
    \mathbf{E}(t)=\int_{t_0}^{t_0+\delta t}\mathbf{e}(t)dt.
\label{eq:eventintegration}
\end{equation}
where $\delta t$ is time interval between two input frames.
The logarithm difference between the last rendered frame and first rendered frame of multiple rendered frames $\{I_i\}_{i=0}^{l}$ gives the estimated event integration maps $\tilde{\mathbf{E}}(t)$.
The event integration loss is then:
\begin{equation}
    L_{event}=\|\mathbf{E}(t)-\widetilde{\mathbf{E}}(t)\|_{1}
\end{equation}
The final loss function is the combination of the blurriness loss and event integration loss as follows:
\begin{equation}
    L=L_{blur}+\lambda_{event}L_{event}
\end{equation}

\section{Experiments}
\label{sec:experiments}

\subsection{Comparisons and Results}
To evaluate the effectiveness of the proposed method, we compare our method with the original Gaussian Splatting (GS), which utilizes blurry images as input and supervision.
Since COLMAP fails with only blurry images, we estimate the initial point cloud using EDI-deblurred images.
We also compare our method with Deblurring 3D Gaussian Splatting(Deblurring-GS) to highlight the advantages of incorporating event data. 

Our evaluation metrics include Peak Signal-to-Noise Ratio (PSNR), Structural Similarity Index (SSIM), and Frames Per Second (FPS). 
All methods are tested on the synthetic data from E\textsuperscript{2}NeRF.
The implementation details are provided in the supplementary materials.

The quantitative comparison results are presented in Table \ref{tab:overall_comparison}. 
\begin{table}[tb]
  \caption{Quantitative analysis. 
  The results in the table are the averages of six synthetic scenes from E\textsuperscript{2}NeRF. 
  We use bold to mark the best result.
  }
  \label{tab:overall_comparison}
  \centering
  \setlength{\tabcolsep}{12pt}
  \begin{tabular}{@{}llll@{}}
    \toprule
    Image Deblur & GS & Deblurring-GS  & Ours\\
    \midrule
    PSNR  & 22.15 & 22.70 & {\bf 29.69}\\
    SSIM & 0.878 & 0.8427 & {\bf 0.9338}\\
    FPS &  190 & 190 & 190\\    
  \bottomrule
  \end{tabular}
\end{table}
The results demonstrate that our method achieves advanced performance compared with other Gaussian Splatting techniques in terms of PSNR and SSIM, while maintaining real-time rendering capabilities with noticeable FPS. 
The advantage of using event data for deblurring RGB images is further evidenced by comparing Deblurring 3D Gaussian Splatting with our method, showing a significant increase in PSNR and SSIM.

\begin{figure}
    \centering
    \includegraphics[width=1.0\textwidth]{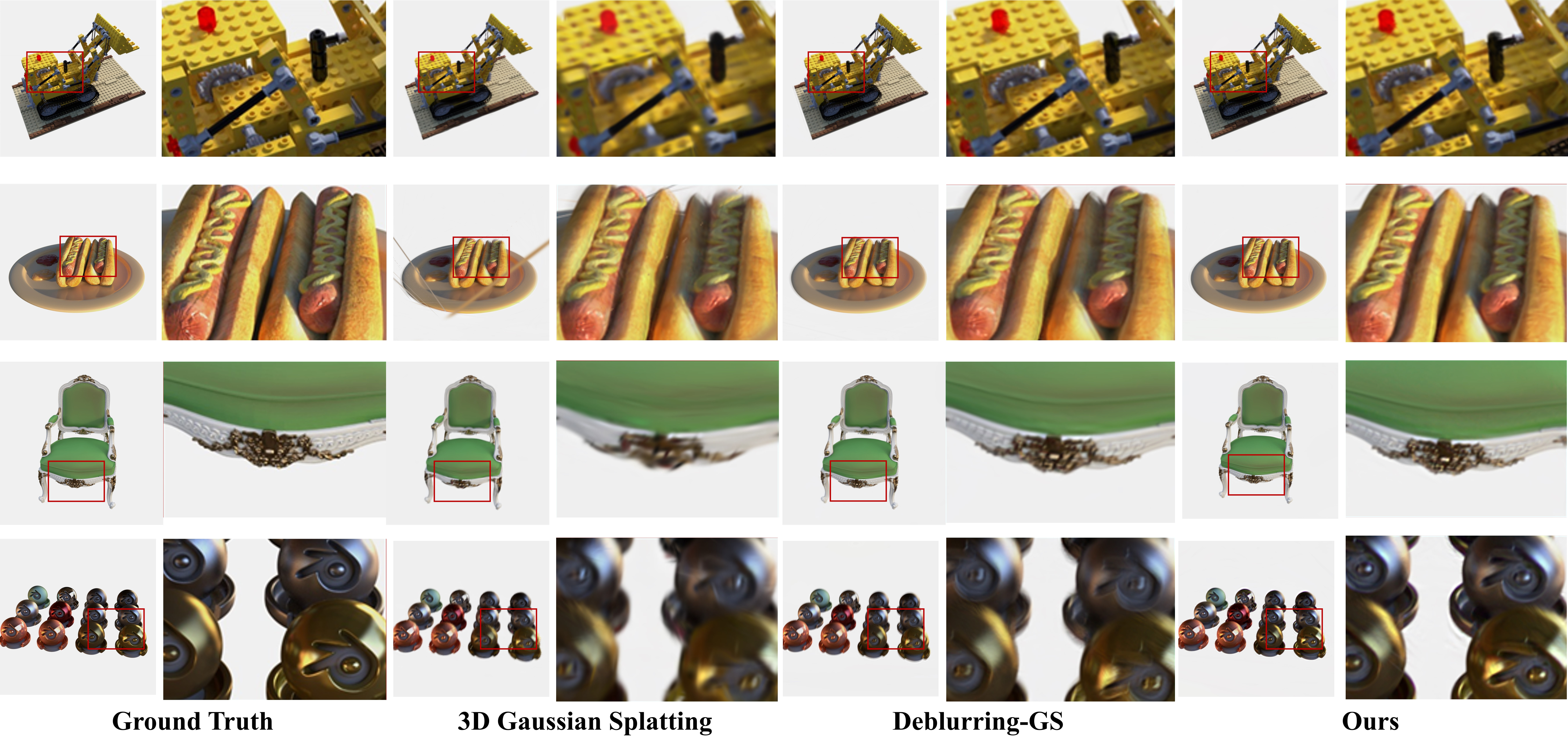}
    \caption{Qualitative results on E\textsuperscript{2}NeRF synthetic dataset.}
    \label{fig:quali-results}
\end{figure}
We also assess the qualitative performance of our method compared to other approaches.
As illustrated in Fig.\ref{fig:quali-results}, the original 3D Gaussian Splatting struggles to produce clear 3D representations in certain scenes, even with EDI-preprocessed relatively sharp input images.
While the Deblurring 3D Gaussian Splatting method can reconstruct relatively sharp objects, it still fails to capture fine details in some cases, such as with the ``materials'' object.
In contrast, our method demonstrates the capability to reconstruct fine-grained, clear objects with greater accuracy.

\subsection{Ablation studies}
\textbf{\textit{Blurriness loss and Event Integration Loss.}}
To validate the effectiveness of training losses, we conduct the experiments on the ``hotdog" scene from E\textsuperscript{2}NeRF dataset.
The results in the Tab.\ref{tab:loss_comparison} demonstrates that both proposed losses significantly improve the deblurring performance, with the event integration loss further slightly enhancing sharp rendering quality.

\begin{table}[tb]
  \caption{Ablation study to validate the effectiveness of losses. 
  "w" represents "with" and "wo" represents "without".
  "$L_{blur}$" denotes the blurriness loss function.
  "$L_{event}$" denotes the event integration loss. 
  }
  \label{tab:loss_comparison}
  \centering
  \setlength{\tabcolsep}{12pt}
  \begin{tabular}{@{}llll@{}}
    \toprule
    Loss Function & wo $L_{blur}$\&$L_{event}$ & w $L_{blur}$ wo$L_{event}$ & w $L_{blur}$\&$L_{event}$ \\
    \midrule
    PSNR  & 22.92 & 31.80 & {\bf 31.83} \\
    SSIM & 0.886 & {\bf 0.9490} & 0.9488 \\ 
  \bottomrule
  \end{tabular}
\end{table}

\noindent\textbf{\textit{With/Without ADE network.}}
We conduct the experiments to further analyze the contribution of ADE module.
Using initial latent poses estimated by COLMAP with EDI preprocessed images, we compare results with and without the ADE network. 
As shown in Table \ref{tab:MLP_comparison}, the ADE module improves all evaluation metrics, indicating its effectiveness in estimating latent poses during exposure.
\begin{table}[tb]
  \caption{Ablation study to analyze the contribution of ADE module. 
  "w" represents "with" and "wo" represents "without".
  }
  \label{tab:MLP_comparison}
  \centering
  \setlength{\tabcolsep}{12pt}
  \begin{tabular}{@{}lll@{}}
    \toprule
    Latent Pose Estimation & wo MLP & w MLP  \\
    \midrule
    PSNR  & 28.76 & {\bf 31.83}  \\
    SSIM  & 0.9347 & {\bf 0.9488}  \\
  \bottomrule
  \end{tabular}
\end{table}
Further ablations are provided in the supplementary materials.

\section{Conclusion}
In this paper, we introduce EaDeblur-GS, which integrates event data and RGB images to achieve sharp neural 3D representations.
We propose an Adaptive Deviation Estimator network to estimate latent poses during exposure by computing Gaussian deviations.
Our method, evaluated against other techniques and through extensive ablation studies, shows significant improvements over the original 3D Gaussian Splatting and Deblurring 3D Gaussian Splatting.

\section*{Acknowledgments}
This work was supported by the National Science and Technology Major Project under Grant 2020AAA0107300, and the Key Research and Development Plan Project of Xinjiang Uygur Autonomous Region under Grant 2023B01006-1.
  

%
%
\newpage
\bibliographystyle{splncs04}
\bibliography{main}

\begin{thebibliography}{10}
\providecommand{\url}[1]{\texttt{#1}}
\providecommand{\urlprefix}{URL }
\providecommand{\doi}[1]{https://doi.org/#1}

\bibitem{kerbl20233d}
Kerbl, B., Kopanas, G., Leimk{\"u}hler, T., Drettakis, G.: 3d gaussian splatting for real-time radiance field rendering. ACM Transactions on Graphics  \textbf{42}(4),  1--14 (2023)

\bibitem{lee2024deblurring}
Lee, B., Lee, H., Sun, X., Ali, U., Park, E.: Deblurring 3d gaussian splatting. arXiv preprint arXiv:2401.00834  (2024)

\bibitem{ma2022deblur}
Ma, L., Li, X., Liao, J., Zhang, Q., Wang, X., Wang, J., Sander, P.V.: Deblur-nerf: Neural radiance fields from blurry images. In: Proceedings of the IEEE/CVF Conference on Computer Vision and Pattern Recognition. pp. 12861--12870 (2022)

\bibitem{mildenhall2021nerf}
Mildenhall, B., Srinivasan, P.P., Tancik, M., Barron, J.T., Ramamoorthi, R., Ng, R.: Nerf: Representing scenes as neural radiance fields for view synthesis. Communications of the ACM  \textbf{65}(1),  99--106 (2021)

\bibitem{pan2019bringing}
Pan, L., Scheerlinck, C., Yu, X., Hartley, R., Liu, M., Dai, Y.: Bringing a blurry frame alive at high frame-rate with an event camera. In: Proceedings of the IEEE/CVF Conference on Computer Vision and Pattern Recognition. pp. 6820--6829 (2019)

\bibitem{qi2023e2nerf}
Qi, Y., Zhu, L., Zhang, Y., Li, J.: E2nerf: Event enhanced neural radiance fields from blurry images. In: Proceedings of the IEEE/CVF International Conference on Computer Vision. pp. 13254--13264 (2023)

\bibitem{rudnev2023eventnerf}
Rudnev, V., Elgharib, M., Theobalt, C., Golyanik, V.: Eventnerf: Neural radiance fields from a single colour event camera. In: Proceedings of the IEEE/CVF Conference on Computer Vision and Pattern Recognition. pp. 4992--5002 (2023)

\bibitem{wang2024mp}
Wang, X., Yin, Z., Zhang, F., Feng, D., Wang, Z.: Mp-nerf: More refined deblurred neural radiance field for 3d reconstruction of blurred images. Knowledge-Based Systems p. 111571 (2024)

\bibitem{Chen_deblurgs2024}
Wenbo, C., Ligang, L.: Deblur-gs: 3d gaussian splatting from camera motion blurred images. Proc. ACM Comput. Graph. Interact. Tech. (Proceedings of I3D 2024)  \textbf{7}(1) (2024). \doi{10.1145/3651301}, \url{http://doi.acm.org/10.1145/3651301}

\bibitem{yu2024evagaussians}
Yu, W., Feng, C., Tang, J., Jia, X., Yuan, L., Tian, Y.: Evagaussians: Event stream assisted gaussian splatting from blurry images. arXiv preprint arXiv:2405.20224  (2024)

\end{thebibliography}
\end{document}